\documentclass{article}

\usepackage[final]{nips_2017}

\usepackage[utf8]{inputenc} 
\usepackage[T1]{fontenc}    
\usepackage{hyperref}       
\usepackage{url}            
\usepackage{booktabs}       
\usepackage{amsfonts}       
\usepackage{nicefrac}       
\usepackage{microtype}      
\usepackage{natbib}
\usepackage{multirow}
\usepackage{graphicx}
\usepackage[table,xcdraw]{xcolor}

\title{Personalization in Goal-oriented Dialog}

\author{
  Chaitanya K. Joshi, Fei Mi, and Boi Faltings \\
  Artificial Intelligence Laboratory \\
  \'Ecole Polytechnique F\'ed\'erale de Lausanne (EPFL) \\
  Lausanne, Switzerland \\
  \texttt{\{firstname.lastname\}@epfl.ch} \\
}
\date{}

\begin{document}

\maketitle

\begin{abstract}
The main goal of modeling human conversation is to create agents which can interact with people in both open-ended and goal-oriented scenarios. End-to-end trained neural dialog systems are an important line of research for such generalized dialog models as they do not resort to any situation-specific handcrafting of rules. However, incorporating personalization into such systems is a largely unexplored topic as there are no existing corpora to facilitate such work. In this paper, we present a new dataset of goal-oriented dialogs which are influenced by speaker profiles attached to them. We analyze the shortcomings of an existing end-to-end dialog system based on Memory Networks and propose modifications to the architecture which enable personalization. We also investigate personalization in dialog as a multi-task learning problem, and show that a single model which shares features among various profiles  outperforms separate models for each profile.
\end{abstract}

\section{Introduction}

The recent advances in memory and attention mechanisms for neural networks architectures have led to remarkable progress in machine translation \citep{JohnsonSLKWCTVW16}, question answering \citep{graves2016hybrid} and other language understanding tasks which require an element of logical reasoning. The main motivation for building neural network based systems over traditional systems for such tasks is that they do not require any feature engineering or domain-specific handcrafting of rules \citep{VinyalsL15}. Conversation modeling is one such domain where end-to-end trained systems have matched or surpassed traditional dialog systems in both open-ended \citep{DodgeGZBCMSW15} and goal-oriented applications \citep{BordesW16}.

An important yet unexplored aspect of dialog systems is the ability to personalize the bot's responses based on the profile or attributes of who it is interacting with. Personalization is key to creating conversational agents that are truly smart and can integrate seamlessly into the lives of human beings. 
For example, a restaurant reservation system should ideally conduct dialog with the user to find values for variables such as location, type of cuisine and price range. It should then make recommendations based on these variables as well as certain fixed attributes about the user (dietary preference, favorite food items, etc.). The register (or style) of the language used by the bot may also be influenced by certain characteristics of the user (age, gender, etc.) \citep{Halliday64}. 
However, there are no open datasets which allow researchers to train end-to-end dialog systems where each conversation is influenced by a speaker's profile \citep{SerbanLCP15}.

With the ultimate aim of creating such a dataset, this paper aims to be an extension of the bAbI dialog dataset introduced by \cite{BordesW16}. Set in the domain of restaurant reservation, their synthetically generated dataset breaks down a conversation into several tasks to test some crucial capabilities that dialog systems should have. Taken together, the tasks can be used as a framework for the analysis of end-to-end dialog systems in a goal-oriented setting. Given a knowledge base (KB) of restaurants and their properties (location, type of cuisine, etc.), the aim of the dialog is to book a restaurant for the user. Full dialogs are divided into various stages, each of which tests if models can learn abilities such as implicit dialog state tracking, using KB facts in dialog, and dealing with new entities not appearing in dialogs from the training set. 

In this paper, we propose extensions to the first five tasks of the existing dataset. In addition to the goal of the original task, the dialog system must leverage a user's profile information to alter speech style and personalize reasoning over the KB. The end-goal is to make a restaurant reservation that is personalized to the user's attributes (dietary preference, favorite food items, etc.).

The synthetic nature of the bAbI dialog dataset and by extension, our work, makes it easy to construct a perfect handcrafted dialog system. Hence, the goal here is not to improve the state of the art in this domain, but to analyze existing end-to-end goal-oriented dialog systems and to model personalization in such frameworks. 
Section 3 presents our modifications to the original dataset and Section 4 describes the various models that are benchmarked on our tasks in Section 5. In Section 6, we analyze the task of conducting goal-oriented dialog as a multi-task learning problem, and show that a single model which shares features among various profiles outperforms separate models for each profile. 

Our dataset is accessible through 
the ParlAI framework \citep{MillerFFLBBPW17} or 
Github along with our experimental code and trained models.\footnote{\texttt{https://github.com/chaitjo/personalized-dialog}}
\section{Related Work}

This work builds upon the bAbI dialog dataset described in \cite{BordesW16}, which is aimed at testing end-to-end dialog systems in the goal-oriented domain of restaurant reservations. Their tasks are meant to complement the bAbI tasks for text understanding and reasoning described in \cite{WestonBCM15}.

The closest work to ours is by \cite{LiGBGD16} who encoded speaker personas into SEQ2SEQ dialog models \citep{VinyalsL15}. The model builds an embedding of a speaker's persona in a vector space based on conversation history of the speaker (for example, all of their tweets). Our work differs from their investigation in the sense that 1) they are concerned with the problem of consistent response generation in an open-ended dialog (chit-chat), whereas our work focuses on the personalization of a goal-oriented conversation and the ranking/discrimination of the correct candidate responses from a set of utterances, and 2) their dialog system needs to be provided with a speaker's conversation history to build the persona, while the bot must compose the user's profile from explicitly provided attributes in our tasks. This is arguably a better representation of real-world learning scenarios where agents can leverage information stored in formal data structures to personalize conversations in domains such as customer care or restaurant reservation.
\section{Dataset Creation}

\begin{figure}
\begin{center}
\includegraphics[width=\linewidth]{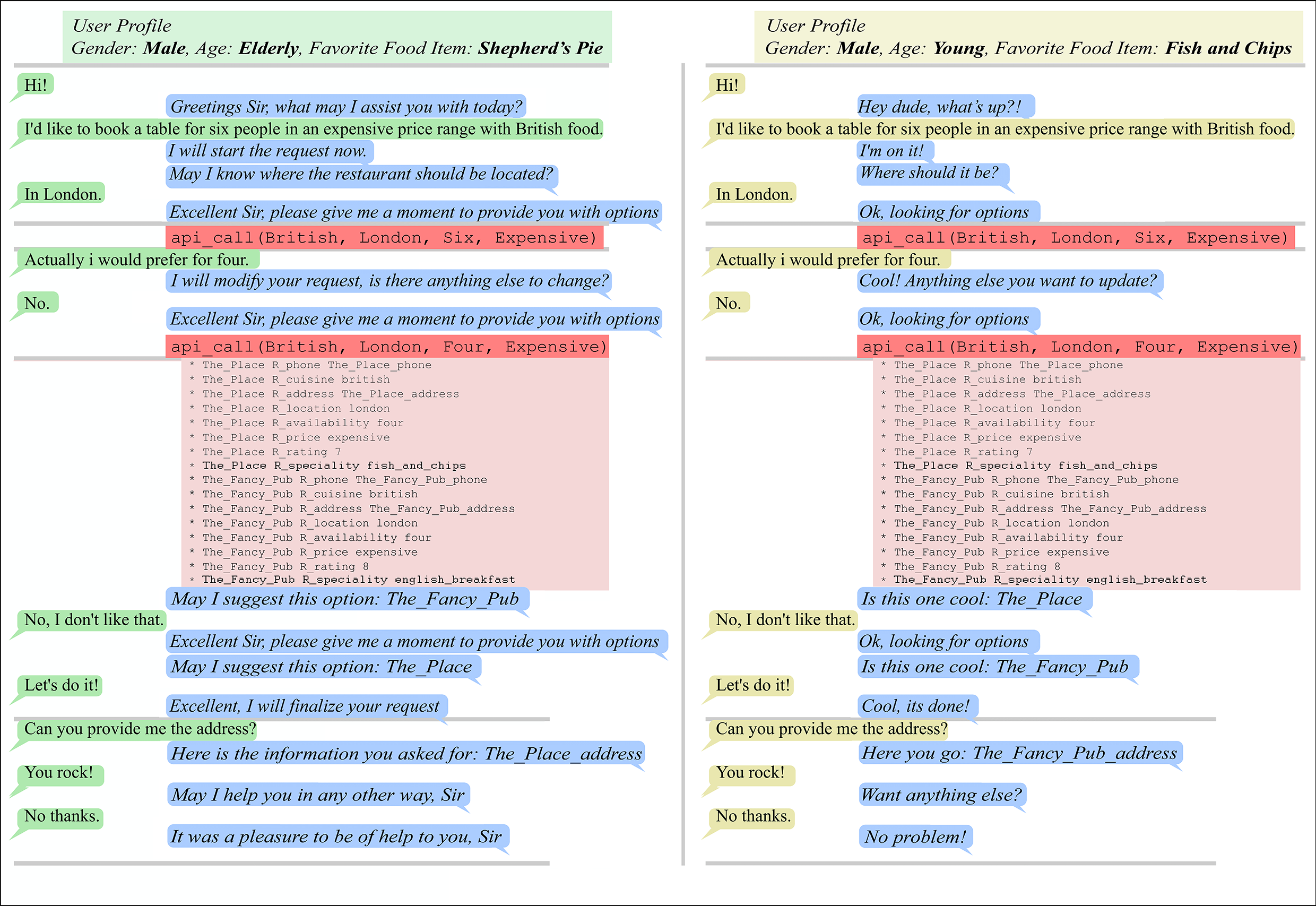}
\caption{\small \textbf{Personalized Restaurant Reservation System.} The user (in green or yellow) conducts a dialog with the bot (in blue) to reserve a table at a restaurant. At each turn, a model has access to the user's profile attributes, the conversation history and the outputs from the API call (in light red) and must predict the next bot utterance or API call (in dark red). The horizontal lines between dialog groups signify the separate tasks that are described in the following sections. (Illustration adapted from Figure 1, \cite{BordesW16}.)}
\label{dialogs}
\end{center}
\end{figure}

We build upon the first five synthetically generated bAbI dialog tasks (T1-T5) where the goal is to book a table at a  restaurant. The conversations are generated by a simulator (in the format shown in Figure \ref{dialogs}) based on an underlying KB containing all the restaurants and their properties. Each restaurant is defined by a type of cuisine (10 choices, e.g., Italian, Indian), a location (10 choices, e.g. London, Tokyo), a price range (cheap, moderate or expensive), a party size (2, 4, 6 or 8 pople) and a rating (from 1 to 8). Each restaurant also has an address and a phone number. Making an API call to the KB returns a list of facts related to all the restaurants that satisfy the four parameters: location, cuisine, price range and party size.

In addition to the user and bot utterances, dialogs in each task are comprised of API calls and the resulting facts. Conversations are generated using natural language patterns after randomly selecting each of the four required fields: location, cuisine, price range and party size. There are 43 patterns for the user and 15 for the bot (the user can say something in up to 4 different ways, while the bot only has one). As described in the following sections, we make further additions to the KB and augment the bot utterance patterns for the creation of our tasks. To fit in with the synthetic nature of the bAbI dialog tasks, the personalization of the bot's responses are handcrafted to be extremely simplistic in comparison to real life situations.

\subsection{Personalized Goal-Oriented Dialog Tasks}
In accordance with the bAbI dialog tasks, tasks 1 and 2 test the model's capability to implicitly track dialog state, tasks 3 and 4 check if they can sort through and use KB facts in conversation. Task 5 combines all aspects of the tasks into a full dialog. 
In addition to fulfilling the original goals of the bAbI dialog tasks, the modified tasks also require the dialog system to personalize its speech style and reasoning over KB facts based on the user's profile, which is composed of various fixed attributes. For each dialog in all tasks, the user's attributes (gender, age, dietary preference and favorite food item) are provided before the first turn of dialog.

\paragraph{Personalization Task 1: Issuing API calls} Users make a query containing from 0 to 4 of the required fields (sampled uniformly). The bot must ask questions to fill the missing fields and then generate the proper API call.

\vspace{-0.1in}
\paragraph{Personalization Task 2: Updating API calls} Starting by issuing an API call as in Task 1, users update their requests between 1 and 4 times (sampled uniformly). The fields to update are selected randomly and the bot must then issue the updated API call.

\vspace{-0.1in}
\paragraph{Personalization Task 3: Displaying Options} Given a user request, the KB is queried by the corresponding API call and the resulting facts are added to the dialog history. The bot must sort the restaurants in the facts using simple heuristics (described in Section 3.3) based on the user's attributes and propose a restaurant to the users until they accept. Users accept a suggestion 25\% of the time or always if it is the last remaining one. 

\vspace{-0.1in}
\paragraph{Personalization Task 4: Providing extra information} Given a user request for a randomly sampled restaurant, all KB facts related to the restaurant are added to the history and the dialog is conducted as if the user has decided to book a table there.  The user then asks for the directions to the restaurant, its contact information or both (with probabilities 25\%, 25\% and 50\% respectively). The bot  must learn to retrieve the correct KB facts from history, tailored for the user.

\vspace{-0.1in}
\paragraph{Personalization Task 5: Conducting full dialogs} Conversations generated for task 5 combine all the aspects of tasks 1-4 into full dialogs. 

\subsection{User Profiles and Speech Style Changes}

The first aspect of personalization incorporated into all 5 tasks was the change in the style of the language used by the bot based on the user's gender (male or female) and age (young, middle-aged or elderly). For each of the 15 bot utterance patterns in the original tasks, we created 6 new patterns for each possible (age, gender) profile permutation. Each of these patterns, while conveying the same information to the user, differed in tone, formality and word usage. Figure \ref{dialogs} illustrates two versions of the same dialog for a (male, elderly) user and a (male, young) user. 

While creating the utterance patterns, importance was given to maintaining a consistent vocabulary for each of the 6 profiles. The levels of formality and precision of the words and language used by the bot increased with age. At the same time, word choices overlapped between the same gender and age group. For example, the pattern for a (male, young) user is similar in formality and tone to the pattern for a (female, young) user and shares certain key words with both (female, young) and (male, middle-aged) user patterns. It is comparatively unrelated to the patterns of a (female, middle-aged) or (female, elderly) user. By creating such relationships between the profiles instead of having 6 completely distinct patterns, we wanted to test whether dialog models could learn to form associations between concepts (such as formality, precision and word usage) and attributes (such as gender and age).

Applying our speech style changes to the bAbI dialog tasks results in 6 versions of the same dialog associated with the 6 user profile. Hence, we increased the size of each task by 6 folds. All the original patterns and the 6 modified patterns associated with each of them are displayed in a tabulated form in Appendix A.

\subsection{KB Updates and Personalized Reasoning}

To personalize the order in which restaurants are recommended by the bot in task 3, we added 2 new attributes to the user's profile: dietary preference (vegetarian or non-vegetarian) and favorite food item (randomly sampled from a list of dishes associated with the cuisine in the dialog). We created a duplicate for each restaurant in the KB, with an additional attribute for type of restaurant (vegetarian or non-vegetarian) to differentiate the otherwise same copies. For every restaurant in the modified KB, we also added the speciality attribute (randomly sampled from a list of dishes associated with the restaurant's cuisine). When modifying each dialog in the original task, instead of proposing the restaurants solely based on their rating (as in the bAbI dialog tasks), we used a score calculated as: rating (out of 8) + 8 (if restaurant type matches user's dietary preference) + 2.5 (if restaurant speciality matches user's favorite food item). With such a metric, all vegetarian restaurants will be scored 8 points higher than non-vegetarian restaurants for a vegetarian user and will always be proposed before the bot suggests any non-vegetarian ones (in descending order of score). By awarding an extra 2.5 points to reflect a user's preference for a favorite food item, a lower rated restaurant specializing in the item will be proposed before a higher rated (by at most 2 points) restaurant specializing in something else. This tests a model's ability to perform true/false reasoning based on the user's profile and to implicitly rank restaurants depending on more than one condition.

Our modification to task 4 requires the bot to retrieve a combination of KB facts related to a restaurant based on certain attributes of the user and the restaurant itself. In addition to the phone number and address, we added 3 new attributes (social media links, parking information and public transport information) to the KB entries for every restaurant. In each modified dialog, when a user asks for the contact information of the restaurant, the bot must return the restaurant's social media link if the user is young, or the phone number if the user is middle-aged or elderly. Similarly, when a user asks for the directions to the restaurant, the bot must return the address and the public transport information if the restaurant is cheap, or the address and the parking information if it is in the moderate or expensive price range. This tests a model's ability to personalize KB fact retrieval based on an attribute in the user's profile (age) or a choice made by the user during the dialog (the restaurant's price range).

\subsection{Updated Dataset}

We generated and structured the dataset in the same way as the original bAbI dialog dataset: for each task, we provided training, validation and test set dialogs generated using half of the modified KB. We also generated another test set from the remaining KB containing new entities (restaurants, cuisine types, etc.) unseen in any training dialog, called Out-Of-Vocabulary (OOV) test set. During training, the model has access to training examples and the KB. Models can be evaluated on both test sets, plain and OOV, on their ability to rank the correct bot utterance at each turn of the dialog from a list of all possible candidates.

The statistics of the datasets for each task are given in Table \ref{data_stat}, along with a comparison to the original bAbI dialog tasks. The size of the vocabulary has increased by almost four folds due to the various speech styles associated with user profiles. The number of possible candidate responses has increased ten fold due to the duplication of each restaurant in the KB and the speech style changes. 
We provide two variations for each task: a full set with all generated dialogs and a small set with only 1000 dialogs each for training, validation and testing to create realistic learning conditions.

\begin{table*}[htb!]
\centering
\caption{\small \textbf{Dataset statistics.} For rows 4 to 6, the first number is the size of the full set and the number in parenthesis is the size of the small set.  (*) PT1-PT5 and all 5 bAbI dialog tasks have two test sets of the same size, one using the same KB entities as the training set and the other using out-of-vocabulary words.
\vspace{0.1in}}
\label{data_stat}
\resizebox{\textwidth}{!}{%
\small
\begin{tabular}{@{}ccccccc@{}}
\toprule
\multirow{2}{*}{\textbf{Task}} & \multirow{2}{*}{\textbf{PT1}} & \multirow{2}{*}{\textbf{PT2}} & \multirow{2}{*}{\textbf{PT3}} & \multirow{2}{*}{\textbf{PT4}} & \multirow{2}{*}{\textbf{PT5}} & \textbf{bAbI dialog}   \\
                               &                               &                               &                               &                               &                               & \textbf{tasks (T1-T5)} \\ \midrule
Vocabulary size                &                               &                               & 14819                         &                               &                               & 3747                   \\
Candidate set size             &                               &                               & 43863                         &                               &                               & 4212                   \\
Training dialogs               & 6000 (1000)                   & 6000 (1000)                   & 12000 (1000)                  & 6000 (1000)                   & 12000 (1000)                  & 1000 each              \\
Validation dialogs             & 6000 (1000)                   & 6000 (1000)                   & 12000 (1000)                  & 6000 (1000)                   & 12000 (1000)                  & 1000 each              \\
Test dialogs                   & 6000* (1000*)                 & 6000* (1000*)                 & 12000* (1000*)                & 6000* (1000*)                 & 12000* (1000*)                & 1000* each             \\ \bottomrule
\end{tabular}
}
\vspace{-0.1in}
\end{table*}
\section{Models}

Following \cite{BordesW16}, we provide baselines on the modified dataset by evaluating several learning methods: rule-based systems, supervised embeddings, and end-to-end Memory networks. We also propose modifications to the Memory Network architecture that facilitate personalized reasoning over KBs.

\subsection{Rule-Based Systems}
Our tasks are generated by modifying and appending to the bAbI dialog tasks T1-T5. All dialogs are built with a rule based simulator and the bot utterances are completely deterministic. Thus, it is possible to create a perfect handcrafted system based on the same rules as the simulator.

\subsection{Supervised Embedding Models}
Although widely known for learning unsupervised embeddings from raw text like in Word2Vec \citep{MikolovCCD13}, embeddings can also be learned in a supervised manner specifically for a given task. Supervised word embedding models which score (conversation history, response) pairs have been shown to be a strong baseline for both open-ended and goal-oriented dialog \citep{DodgeGZBCMSW15,BordesW16}. We do not handcraft any special embeddings for the user profiles and treat it as a turn in the dialog. 

The embedding vectors are trained specifically for the task of predicting the next response given the previous conversation: a candidate response $y$ is scored against the input $x$: $f(x,y) = (Ax)^\top By$, where $A$ and $B$ are $d \times V$ word embedding matrices and inputs and responses are summed bag-of-embeddings. The model is trained with SGD to minimize a margin ranking loss: $f(x,y) > m + f(x,\bar{y})$, where $m$ is the size of the margin and $N$ {\em negative} candidate responses $\bar{y}$ are sampled per example.

\subsection{Memory Networks}
Memory Networks \citep{WestonCB14} are a recent class of models that have proven successful for a variety of tasks such as question answering \citep{SukhbaatarSWF15} and conducting dialog \citep{DodgeGZBCMSW15}. For dialogs, the entire conversation history is stored in the memory component of the model. It can be iteratively read from to perform reasoning and select the best possible responses based on the context. The model's attention to various entries in memory at each turn of conversation can be visualized over multiple iterations (called hops). Implementing the modifications to the Memory Network architecture described by \cite{BordesW16}, we use the model as an end-to-end baseline and analyze its performance. 

The user profile information is stored in the memory of the model as if it were the first turn of the conversation history spoken by the user, i.e. the model builds an embedding of the profile by combining the values of the embeddings of each attribute in the profile.  Unlike \cite{BordesW16}, we do not make use of any match type features for KB entities. Our goal is to analyze the capabilities of the existing Memory Network model to leverage profile information when conducting dialog. Appendix B contains visualizations of Memory Network predictions based on the experiments described in the following section.

\subsection{\textit{Split Memory} Architecture}
The personalized reasoning aspect of Tasks 3 and 4 require the bot to combine information from the user's attributes as well as the conversation history (i.e. KB facts and dialogs). A single memory architecture is not be able to focus sufficiently on the various attributes and dialog turns used to formulate a personalized response. 

\begin{figure}[htb!]
\begin{center}
\includegraphics[width=\linewidth]{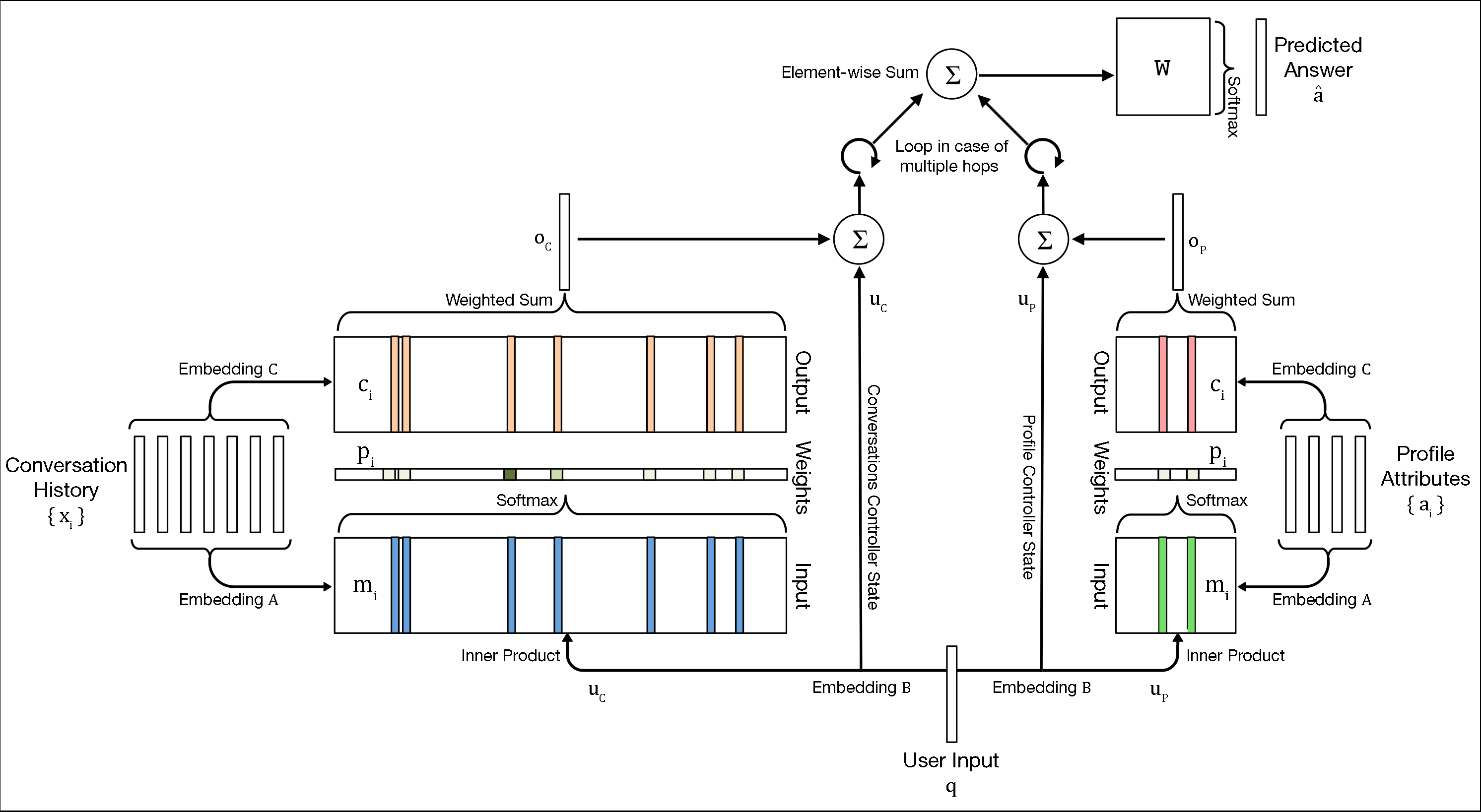}
\caption{\small \textbf{\textit{Split memory} architecture for Memory Networks.} Profile attributes and conversation history are modeled in two separate memories. The outputs from both memories are summed to get the final response.}
\label{memnn-split}
\end{center}
\end{figure}

We propose dividing the memory of the model into two halves: profile attributes and conversation history. The various attributes are added as separate entries in the profile memory before the dialog starts. At the same time, each dialog turn is added to the conversation memory as the dialog progresses. The mechanism through which the model reads from and reasons over each memory remains the same. The outputs from both memories are summed element-wise to get the final response for each conversation turn. A visualization of the model is shown in Figure \ref{memnn-split}. 

\section{Experiments}

We report per-response accuracy (i.e. the percentage of responses in which the correct candidate is chosen out of all possible ones) across all the models and tasks in Table \ref{results}. The rows show tasks PT1-PT5 and columns 2-5 give the accuracy for each of the models. The hyperparameters for the models were optimized on the validation sets (values are provided in Appendix C).

\begin{table}[htb!]
	\centering
	\caption{\small \textbf{Test results across models and tasks.} For Memory Networks, the first number is the accuracy on the full set of dialogs for each task and the number in parenthesis is the accuracy on the small set (with 1000 dialogs). All other models were evaluated on the full set only.
    \vspace{0.1in}}
	\label{results}
	\small
	\begin{tabular}{@{}lcccc@{}}
		\toprule
		\textbf{} & \textbf{Rule-based} & \textbf{Supervised} & \multicolumn{2}{c}{\textbf{Memory Networks}} \\
		\textbf{Task}  & \textbf{System} & \textbf{Embeddings} & \textbf{Standard} & \textbf{Split memory} \\ \midrule
		PT1: Issuing API calls & 100 & 84.37 & 99.83 (98.87) & 85.66 (82.44) \\
		PT2: Updating API calls & 100 & 12.07 & 99.99 (99.93) & 93.42 (91.27) \\
		PT3: Displaying options & 100 & 9.21 & 58.94 (58.71) & 68.60 (68.56) \\
		PT4: Providing information & 100 & 4.76 & 57.17 (57.17) & 57.17 (57.11) \\
		PT5: Full dialog & 100 & 51.6 & 85.10 (77.74) & 87.28 (78.1) \\ \bottomrule
	\end{tabular}
\end{table}

\paragraph{Handcrafted systems} As expected, handcrafted rule-based systems outperformed all machine learning models and solved all 5 tasks perfectly. However, it is important to note that building a rule-based system for real conversations is not easy: our tasks use a restricted vocabulary and fixed speech patterns. As such, we want to use them to analyze the strengths and weaknesses of machine learning algorithms.

\paragraph{Supervised Embeddings} Compared to results reported on the bAbI dialog tasks in \cite{BordesW16}, supervised embeddings performed significantly worse on the modified tasks. The model was unable to complete any of the tasks successfully and had extremely low per-response accuracy for PT2-PT5. We attribute this drop in performance to the increased complexity of our tasks due to the four fold increase in vocabulary and the ten fold increase in candidate set size.

\paragraph{Memory Networks for speech style} Memory Networks substantially outperformed supervised embeddings for all tasks. They completed PT1 and PT2 (issuing and updating API calls) with a very high degree of accuracy. This indicates that the model is able to implicitly track dialog state and personalize the bot's utterance based on the user's profile. Tables \ref{memnn-pt1} to \ref{memnn-pt2} show visualizations of the parts of the memory read from at each turn of the dialog. 

\paragraph{\textit{Split memory} model for personalized reasoning} Results on PT3-PT5 suggest that Memory Networks were unable to use KB facts and profile information in conversation reliably. Memory Networks with the \textit{split memory} feature significantly outperformed the standard architecture for PT3, which lead to better performance on the full dialog task PT5. Visualization and analysis of the model's attention weights in Tables \ref{memnn-pt3} to \ref{memnn-sm-pt4} shows that standard Memory Networks fail to interpret knowledge about entities and link it to the user's attributes, while splitting memory helps to combine profile information with the correct KB facts when sorting restaurants (in PT3) or providing additional information about them (in PT4).

\paragraph{Difficulty in training} Despite extensive hyperparameter tuning, the \textit{split memory} model proved hard to train for the PT1 and PT2. Training converged at significantly lower accuracy than the standard model, resulting in lower test accuracy and suggesting that a simpler model is more suitable for tasks which do not require compositional reasoning over various entries in the memory. In general, learning distinguishable embeddings for each entry in the KB proved to be a challenge for all models.

\paragraph{Generic Architecture for Personalization} Although no model was able to solve PT3 and PT4 sufficiently, results with the \textit{split memory} architecture for Memory Networks are encouraging. We believe that such a modification is not specific to Memory Networks and can be applied to any generic dialog model which stores and reasons over conversation turns.


\section{Multi-task Learning}

We analyzed the Memory Network architecture in a multi-task learning scenario for conducting full dialog. We trained individual profile-specific models for each of the 6 profile permutations for speech style changes and compared their performance to a single multi-profile model. Each of the profile-specific models were trained on 1000 full dialogs between the bot and a user with the corresponding age and gender combination (6000 dialogs in total). The multi-profile model was trained on the same 6000 full dialogs, containing all 6 user profiles. For each profile, we report per-response accuracies for both the profile-specific and multi-profile model on 1000 test dialogs (with users having the same profile) in Table \ref{mtl-results}. 

The multi-profile model significantly outperformed each profile-specific model, indicating that training a single model on dialogs with multiple profiles which share logic and vocabulary is an effective learning strategy. We attribute the bump in accuracy to learning shared features among the 6 user profiles, which can be illustrated by comparing the attention weights for the two strategies in Table \ref{memnn-mtl}.

\begin{table}[htb!]
	\centering
	\caption{\small \textbf{Test results for multi-task learning scenarios.}
    \vspace{0.1in}}
	\label{mtl-results}
	\small
	\begin{tabular}{@{}ccc@{}}
		\toprule
		\textbf{Profile} & \textbf{Profile-specific model} & \textbf{Multi-profile model} \\ \midrule
		male, young & 80.38 & 85.04 \\
		female, young & 80.15 & 84.91 \\
		male, middle-aged & 80.29 & 84.71 \\
		female, middle-aged & 80.21 & 85.14 \\
		male, elderly & 80.57 & 85.46 \\
		female, elderly & 80.41 & 84.95 \\ \bottomrule
	\end{tabular}
\end{table}
\section{Conclusion}

This paper aims to bridge a gap in research on neural conversational agents by introducing a new dataset of goal-oriented dialogs with user profiles associated with each dialog. The dataset acts as a testbed for the training and analysis of end-to-end goal-oriented conversational agents which must personalize their conversation with the user based on attributes in the user's profile. Crucial aspects of goal-oriented conversation have been split into various synthetically generated tasks to evaluate the strengths and weaknesses of models in a systematic way before applying them on real data. Despite the language and scenarios of the tasks being artificial, we believe that building mechanisms that can solve them is a reasonable starting point towards the development of sophisticated personalized dialog systems in domains such as restaurant reservation, customer care or digital assistants.

Additionally, we demonstrated how to use our tasks to break down a dialog system based on Memory Networks. The model was unable to perform compositional reasoning or personalization to solve the tasks, indicating that further work needs to be done on learning methods for these aspects. Towards this goal, we proposed a \textit{split memory} architecture for end-to-end trained dialog agents that lead to an improvement in overall performance.

Finally, we demonstrated the advantages of formulating personalization in dialog as a multi-task learning problem as opposed to training separate models for various user profile. We showed that a jointly trained model significantly outperformed profile-specific models and is able to leverage shared features and relationships between the various conversation styles.

\newpage

\bibliographystyle{plainnat}
\begin{small}
\bibliography{main.bib}
\end{small}

\newpage

\section*{A. Speech Style Changes}
\label{ssc}
Table \ref{ssc-table} displays the 15 original bot utterance patterns from the bAbI dialog tasks and the 6 profile-based modified utterance patterns associated with each of them. Column 1 shows the original bot utterance patterns from the bAbI dialog tasks. Columns 2-7 show the modified utterance patterns associated with each of the 6 profiles.

\section*{B. Examples of Predictions of a Memory Network}
\label{memnn-viz}
Tables \ref{memnn-pt1}-\ref{memnn-sm-pt4} illustrate examples of predictions by Memory Networks to support the various tasks and experiments described in the paper. All models were trained on the full dataset.

At any turn of the dialog, the Memory Network stores the conversation history in its memory and, based on the user’s input for that turn, pays attention to specific utterances from the memory. It can iteratively reason over the memory and uses a weighted combination of these utterances to predict the bot’s response to the user. In our visualization, we take the model state at a specific turn in the conversation and highlight the values of the attention weights over the memory for each iteration (called a hop). In case of the \textit{split memory} architecture, we show the attention weights over the profile memory separately.

\section*{C. Hyperparameters}
Tables \ref{se-hp} and \ref{se-memnn} display the hyperparameters used to train the best performing models for each task.

\newpage

\begin{table}[]
\centering
\caption{\small \textbf{Speech style changes based on profile.} Ellipses have been used in place of various KB entries such as restaurant names or attributes.
\vspace{0.1in}}
\label{ssc-table}
\resizebox{\textwidth}{!}{%
\begin{tabular}{@{}lllllll@{}}
\toprule
\textbf{} & \multicolumn{6}{c}{\textbf{Modified bot utterance}} \\
\textbf{Original bot} & \multicolumn{1}{c}{\textbf{male,}} & \multicolumn{1}{c}{\textbf{female,}} & \multicolumn{1}{c}{\textbf{male,}} & \multicolumn{1}{c}{\textbf{female,}} & \multicolumn{1}{c}{\textbf{male,}} & \multicolumn{1}{c}{\textbf{female,}} \\
\textbf{utterance} & \multicolumn{1}{c}{\textbf{young}} & \multicolumn{1}{c}{\textbf{young}} & \multicolumn{1}{c}{\textbf{middle-aged}} & \multicolumn{1}{c}{\textbf{middle-aged}} & \multicolumn{1}{c}{\textbf{elderly}} & \multicolumn{1}{c}{\textbf{elderly}} \\ \midrule
hello what can & hey dude what & hey girl how & hello sir what & hello maam & greetings sir what & good day madam \\
i help you & is up & is it going & can i help & how can i & may i assist & how could i \\
with today &  &  & you with & help you & you with today & assist you today \\ \midrule
 & i'm on & be right back & i'm processing & give me a second & excellent sir & thank you madam \\
i'm on it & your request & with your & the request & for processing & i will start & i shall start the \\
 &  & reservation &  & the reservation & the request now & reservation now \\ \midrule
any preference & what food & what food & what type of & what type of & may i know your & could you tell me \\
on a type & are you & are you & cuisine would & cuisine would & preference on the & your preference on \\
of cuisine & looking for & looking for & you like to eat & you like to eat & type of cusine & the type of cusine \\ \midrule
where should & where should & where should & where should & where should & may i know & could you tell me \\
it be & it be & it be & it be & it be & where the restaurant & where the restaurant \\
 &  &  & located & located & should be located & should be located \\ \midrule
which price & what should & what should & which price & which price & may i know & would you mind \\
range are & the price be & the price be & range are you & range are you & your prefered & telling me your \\
looking for &  &  & looking for & looking for & price range & price range \\ \midrule
how many people & how many & how many & how many people & how many people & may i know how & would you mind telling \\
would be in & are you & are you & would be in & would be in & many guests will & me how may guests \\
your party &  &  & your party & your party & be at your table & shall be at your table \\ \midrule
api\_call ... & api\_call ... & api\_call ... & api\_call ... & api\_call ... & api\_call ... & api\_call ... \\ \midrule
sure is there & cool anything & awesome is & great is there & great is there & i will modify your & i shall modify \\
anything else & else you want & there any & anything else & anything else & request is there & your reservation \\
to update & to update & other update & to modify & to modify & anything else & is there any \\
 &  &  &  &  & to change & other change \\ \midrule
ok let me & ok looking & sure finding & ok sir i'm & sure maam i'm & excellent sir please & thank you madam \\
look into some & for options & some options & looking for & finding some & give me a moment & i shall provide \\
options for you &  &  & options for & options for & to provide you & you with options \\
 &  &  & you & you & with options & shortly \\ \midrule
what do you & is this one & how about & is this & what do you & may i & would you \\
think of this & cool: ... & this one: ... & a good & think of this & suggest this & consider this \\
option: ... &  &  & option: ... & option: ... & option: ... & option: ... \\ \midrule
sure let me & ok looking & sure finding & ok i'll look & sure i'll find & yes sir i will & yes maam i shall \\
find an other & for something & something & for a better & a better & look for another & find another \\
option for you & else & else & option & option & suitable option & suitable option \\ \midrule
here it & here you & here you & here it & here it & here is the & here is the \\
is: ... & go ... & go ... & is ... & is ... & information & information \\
 &  &  &  &  & you asked for ... & you asked for ... \\ \midrule
is there anything & want & need & is there & can i assist you & may i help & could i assist you \\
i can help & anything & something & anything i can & with something & you in any & in some other \\
you with & else & else & help you with & else & other way sir & manner madam \\ \midrule
great let me & cool its & awesome & great i'll & great let me & excellent i will & thank you i \\
do the & done & you are & finalize the & do the & finalize your & shall finish your \\
reservation &  & done & request & reservation & request & reservation \\ \midrule
you're & no & happy to & you're & you're & it was a pleasure & i am grateful \\
welcome & problem & help & welcome & welcome & to be of help & to assist you \\
 &  &  &  &  & to you sir & madam \\ \bottomrule
\end{tabular}%
}
\end{table}


\begin{table}[]
\centering
\caption{\small \textbf{Personalization task 1 (Issuing API calls).} The model solves the task successfully by directing its attention towards the memories containing the values of the fields of the API call. The signal grows stronger on the second hop and switches almost exclusively to the price range in the third. This indicates that the model does not need multiple hops for successfully completing PT1. (Zero attention weight signifies a very small value.)
\vspace{0.1in}}
\label{memnn-pt1}
\resizebox{\textwidth}{!}{%
\begin{tabular}{|l|l|l|c|c|c|}
\hline
\textbf{Time} & \textbf{Locutor} & \textbf{Dialog History} & \textbf{Hop \#1} & \textbf{Hop \#2} & \textbf{Hop \#3} \\ \hline
1 &  & female elderly & 0.006 & 0 & 0 \\
2 & User & hi & 0.012 & 0 & 0 \\
3 & Bot & good day madam how could i assist you today & 0 & 0 & 0 \\
4 & User & may i have a table & 0.006 & \cellcolor[HTML]{FD6864}0.248 & 0 \\
5 & Bot & thank you madam i shall start the reservation now & 0 & 0 & 0 \\
6 & User & <SILENCE> & \cellcolor[HTML]{FFCCC9}0.064 & 0 & 0 \\
7 & Bot & could you tell me your preference on the type of cuisine & 0.001 & 0 & 0 \\
8 & User & i love french food & \cellcolor[HTML]{FD6864}0.307 & \cellcolor[HTML]{FFCCC9}0.103 & 0 \\
9 & Bot & could you tell me where the restaurant should be located & 0 & 0 & 0 \\
10 & User & madrid please & \cellcolor[HTML]{FD6864}0.265 & \cellcolor[HTML]{FD6864}0.282 & 0 \\
11 & Bot & would you mind telling me how many guests shall be at your table & 0 & 0 & 0 \\
12 & User & for four please & \cellcolor[HTML]{FD6864}0.225 & \cellcolor[HTML]{FFCCC9}0.076 & 0 \\
13 & Bot & would you mind telling me your price range & 0 & 0 & 0 \\
14 & User & in a cheap price range please & \cellcolor[HTML]{FFCCC9}0.11 & \cellcolor[HTML]{FD6864}0.29 & \cellcolor[HTML]{FE0000}0.999 \\
15 & Bot & thank you madam i shall provide you with options shortly & 0 & 0 & 0 \\ \hline
\multicolumn{2}{|l|}{\textbf{User input}} & \multicolumn{4}{l|}{<SILENCE>} \\ \hline
\multicolumn{2}{|l|}{\textbf{Correct answer}} & \multicolumn{4}{l|}{api\_call french madrid four cheap} \\ \hline
\multicolumn{2}{|l|}{\textbf{Predicted answer}} & \multicolumn{4}{l|}{api\_call french madrid four cheap} \\ \hline
\end{tabular}%
}
\end{table}

\begin{table}[]
\centering
\caption{\small \textbf{Personalization task 2 (Updating API calls).} The model correctly learns to focus on the original API call and the changes demanded by the user. It does not require multiple hops to solve this task.
\vspace{0.1in}}
\label{memnn-pt2}
\resizebox{\textwidth}{!}{%
\begin{tabular}{|l|l|l|c|c|c|}
\hline
\textbf{Time} & \textbf{Locutor} & \textbf{Dialog History} & \textbf{Hop \#1} & \textbf{Hop \#2} & \textbf{Hop \#3} \\ \hline
1 &  & female middle-aged & 0.004 & 0 & 0 \\
2 & User & good morning & 0.008 & 0 & 0 \\
3 & Bot & hello maam how can i help you & 0 & 0 & 0 \\
4 & User & may i have a table in bombay with spanish food for four in a expensive price range & 0.001 & 0 & \cellcolor[HTML]{FE0000}0.999 \\
5 & Bot & give me a second for processing the reservation & 0.001 & 0 & 0 \\
6 & User & <SILENCE> & \cellcolor[HTML]{FFCCC9}0.082 & 0 & 0 \\
7 & Bot & sure maam i'm finding some options for you & 0 & 0 & 0 \\
8 & User & <SILENCE> & 0.05 & 0 & 0 \\
9 & Bot & api\_call spanish bombay four expensive & \cellcolor[HTML]{FFCCC9}0.214 & 0.002 & 0 \\
10 & User & actually i would prefer with french food & \cellcolor[HTML]{FD6864}0.635 & \cellcolor[HTML]{FE0000}0.997 & 0 \\
11 & Bot & great is there any other thing to modify & 0 & 0 & 0 \\
12 & User & no & 0.001 & 0 & 0 \\
13 & Bot & sure maam i'm finding some options for you & 0 & 0 & 0 \\ \hline
\multicolumn{2}{|l|}{\textbf{User input}} & \multicolumn{4}{l|}{<SILENCE>} \\ \hline
\multicolumn{2}{|l|}{\textbf{Correct answer}} & \multicolumn{4}{l|}{api\_call french bombay four expensive} \\ \hline
\multicolumn{2}{|l|}{\textbf{Predicted answer}} & \multicolumn{4}{l|}{api\_call french bombay four expensive} \\ \hline
\end{tabular}%
}
\end{table}

\begin{table}[]
\centering
\caption{\small \textbf{Speech style and user profiles.} In order to study how the model uses profiles to modify speech style, we analyzed a turn of the dialog unrelated to any tasks. The model learns to focus on the user’s profile along with the utterance containing the incomplete demand for the table. However, it also pays attention to the salutation, which intuitively should not have any impact on the bot’s output at the given turn. Multiple hops are unnecessary for modeling speech style.
\vspace{0.1in}}
\label{memnn-ssc}
\resizebox{\textwidth}{!}{%
\begin{tabular}{|l|l|l|c|c|c|}
\hline
\textbf{Time} & \textbf{Locutor} & \textbf{Dialog History} & \textbf{Hop \#1} & \textbf{Hop \#2} & \textbf{Hop \#3} \\ \hline
1 &  & male elderly & \cellcolor[HTML]{FFCCC9}0.15 & 0 & 0 \\
2 & User & hello & \cellcolor[HTML]{FD6864}0.306 & 0 & 0 \\
3 & Bot & greetings sir what may i assist you with today & 0.008 & 0 & 0 \\
4 & User & may i have a table in a moderate price range with italian food for eight & \cellcolor[HTML]{FD6864}0.536 & \cellcolor[HTML]{FE0000}0.999 & 0 \\
5 & Bot & excellent sir i will start the request now & 0 & 0 & \cellcolor[HTML]{FE0000}0.999 \\ \hline
\multicolumn{2}{|l|}{\textbf{User input}} & \multicolumn{4}{l|}{<SILENCE>} \\ \hline
\multicolumn{2}{|l|}{\textbf{Correct answer}} & \multicolumn{4}{l|}{may i know where the restaurant should be located} \\ \hline
\multicolumn{2}{|l|}{\textbf{Predicted answer}} & \multicolumn{4}{l|}{may i know where the restaurant should be located} \\ \hline
\end{tabular}%
}
\end{table}

\begin{table}[]
\centering
\caption{\small \textbf{Personalization task 3 (Displaying options).} The model should ideally be focusing on factors that are used for implicit ranking, such as the user’s profile and the ratings, types and specialities of the various restaurants in the KB facts. It should also pay attention to the restaurants that have already been suggested to the user. However, it attends primarily to the locations, indicating that it is insufficient at reasoning over the KB. We have only shown important utterances in the table.
\vspace{0.1in}}
\label{memnn-pt3}
\resizebox{\textwidth}{!}{%
\begin{tabular}{|l|l|l|c|c|c|}
\hline
\textbf{Time} & \textbf{Locutor} & \textbf{Dialog History} & \textbf{Hop \#1} & \textbf{Hop \#2} & \textbf{Hop \#3} \\ \hline
1 &  & female young non-veg pizza & \cellcolor[HTML]{FFCCC9}0.0001 & 0 & 0 \\
20 & User & resto\_rome\_moderate\_italian\_8stars\_1 R\_rating 8 & \cellcolor[HTML]{FFCCC9}0.0001 & 0 & 0 \\
21 & User & resto\_rome\_moderate\_italian\_8stars\_1 R\_type veg & 0 & 0 & 0 \\
22 & User & resto\_rome\_moderate\_italian\_8stars\_1 R\_speciality pizza & 0 & 0 & 0 \\
29 & User & resto\_rome\_moderate\_italian\_3stars\_1 R\_location rome & \cellcolor[HTML]{FE0000}0.879 & \cellcolor[HTML]{FE0000}0.999 & 0 \\
33 & User & resto\_rome\_moderate\_italian\_3stars\_1 R\_type veg & \cellcolor[HTML]{FFCCC9}0.0001 & 0 & 0 \\
44 & User & resto\_rome\_moderate\_italian\_2stars\_2 R\_rating 2 & 0 & 0 & 0 \\
45 & User & resto\_rome\_moderate\_italian\_2stars\_2 R\_type non-veg & \cellcolor[HTML]{FFCCC9}0.0001 & 0 & 0 \\
46 & User & resto\_rome\_moderate\_italian\_2stars\_2 R\_speciality pasta & 0 & 0 & 0 \\
74 & User & hi & 0 & 0 & 0 \\
75 & Bot & hey girl how is it going & 0 & 0 & 0 \\
76 & User & may i have a table in rome with italian food in a moderate price range & 0 & 0 & 0 \\
77 & Bot & be right back with your reservation & 0 & 0 & \cellcolor[HTML]{FE0000}0.999 \\
83 & Bot & how about this one: resto\_rome\_moderate\_italian\_8stars\_2 & 0 & 0 & 0 \\
84 & User & no i don't like that & 0 & 0 & 0 \\
85 & Bot & sure finding something else & 0 & 0 & 0 \\
86 & User & <SILENCE> & \cellcolor[HTML]{FFCCC9}0.0003 & 0 & 0 \\
87 & Bot & how about this one: resto\_rome\_moderate\_italian\_3stars\_2 & 0 & 0 & 0 \\
88 & User & no i don't like that & 0 & 0 & 0 \\
89 & Bot & sure finding something else & 0 & 0 & 0 \\ \hline
\multicolumn{2}{|l|}{\textbf{User Input}} & \multicolumn{4}{l|}{<SILENCE>} \\ \hline
\multicolumn{2}{|l|}{\textbf{Correct Answer}} & \multicolumn{4}{l|}{how about this one: resto\_rome\_moderate\_italian\_8stars\_1} \\ \hline
\multicolumn{2}{|l|}{\textbf{Predicted Answer}} & \multicolumn{4}{l|}{what food are you looking for} \\ \hline
\end{tabular}%
}
\end{table}

\begin{table}[]
\centering
\caption{\small \textbf{Personalization task 3 with \textit{split memory} architecture.} The model is unable to focus on the ratings, types and specialities of the various restaurants. However, it pays extra attention to the restaurants that have already been suggested. It also attends to the user's profile in a meaningful way- it iterates over the dietary preferences and favorite food item to perform reasoning, and then over the gender and age to modify speech style. The predicted response is more meaningful than the standard Memory Network, although it is unable to distinguish between the embeddings of various restaurants.
\vspace{0.1in}}
\label{memnn-sm-pt3}
\resizebox{\textwidth}{!}{%
\begin{tabular}{ll|l|c|c|c|}
\cline{3-6}
                                           &                        & \multicolumn{1}{c|}{\textbf{Profile}}                                   & \textbf{Hop \#1}                                 & \textbf{Hop \#2}               & \textbf{Hop \#3}              \\ \cline{3-6} 
                                           &                        & \multicolumn{1}{c|}{female}                                             & 0.011                                            & \cellcolor[HTML]{FD6864}0.571  & 0                             \\
                                           &                        & \multicolumn{1}{c|}{young}                                              & 0.017                                            & \cellcolor[HTML]{FFCCC9}0.423  & 0                             \\
                                           &                        & \multicolumn{1}{c|}{non-veg}                                            & \cellcolor[HTML]{FFCCC9}0.442                    & 0.006                          & \cellcolor[HTML]{FE0000}0.999 \\
                                           &                        & \multicolumn{1}{c|}{pizza}                                              & \cellcolor[HTML]{FD6864}0.53                     & 0                              & 0                             \\ \cline{1-3}
\multicolumn{1}{|l|}{\textbf{Time}}        & \textbf{Locutor}       & \textbf{Dialog History}                                                 &                                                  &                                &                               \\ \cline{1-3}
\multicolumn{1}{|l|}{19}                   & User                   & resto\_rome\_moderate\_italian\_8stars\_1 R\_rating 8                   & 0.0001                                           & 0                              & 0                             \\
\multicolumn{1}{|l|}{20}                   & User                   & resto\_rome\_moderate\_italian\_8stars\_1 R\_type veg                   & 0                                                & 0                              & 0                             \\
\multicolumn{1}{|l|}{21}                   & User                   & resto\_rome\_moderate\_italian\_8stars\_1 R\_speciality pizza           & 0.0002                                           & 0                              & 0                             \\
\multicolumn{1}{|l|}{}                     &                        & resto\_rome\_moderate\_italian\_8stars\_1 R\_parking                    & \cellcolor[HTML]{FFCCC9}                         &                                &                               \\
\multicolumn{1}{|l|}{\multirow{-2}{*}{23}} & \multirow{-2}{*}{User} & \multicolumn{1}{c|}{resto\_rome\_moderate\_italian\_8stars\_1\_parking} & \multirow{-2}{*}{\cellcolor[HTML]{FFCCC9}0.0127} & \multirow{-2}{*}{0}            & \multirow{-2}{*}{0}           \\
\multicolumn{1}{|l|}{31}                   & User                   & resto\_rome\_moderate\_italian\_3stars\_1 R\_rating 3                   & 0                                                & 0                              & 0                             \\
\multicolumn{1}{|l|}{43}                   & User                   & resto\_rome\_moderate\_italian\_2stars\_2 R\_rating 2                   & 0                                                & 0                              & 0                             \\
\multicolumn{1}{|l|}{44}                   & User                   & resto\_rome\_moderate\_italian\_2stars\_2 R\_type non-veg               & 0                                                & 0                              & 0                             \\
\multicolumn{1}{|l|}{45}                   & User                   & resto\_rome\_moderate\_italian\_2stars\_2 R\_speciality pasta           & 0                                                & 0                              & 0                             \\
\multicolumn{1}{|l|}{}                     &                        & resto\_rome\_moderate\_italian\_2stars\_1 R\_address                    & \cellcolor[HTML]{FFCCC9}                         &                                &                               \\
\multicolumn{1}{|l|}{\multirow{-2}{*}{51}} & \multirow{-2}{*}{User} & \multicolumn{1}{c|}{resto\_rome\_moderate\_italian\_2stars\_1\_address} & \multirow{-2}{*}{\cellcolor[HTML]{FFCCC9}0.0184} & \multirow{-2}{*}{0}            & \multirow{-2}{*}{0}           \\
\multicolumn{1}{|l|}{73}                   & User                   & hi                                                                      & 0.0002                                           & 0                              & 0                             \\
\multicolumn{1}{|l|}{74}                   & Bot                    & hey girl how is it going                                                & 0.0006                                           & 0                              & 0                             \\
\multicolumn{1}{|l|}{75}                   & User                   & may i have a table in rome with italian food in a moderate price range  & 0.0022                                           & 0                              & 0                             \\
\multicolumn{1}{|l|}{80}                   & Bot                    & sure finding some options                                               & 0.0002                                           & 0.0002                         & \cellcolor[HTML]{FE0000}0.999 \\
\multicolumn{1}{|l|}{81}                   & User                   & <SILENCE>                                            & 0.0008                                           & 0                              & 0                             \\
\multicolumn{1}{|l|}{82}                   & Bot                    & how about this one: resto\_rome\_moderate\_italian\_8stars\_2           & \cellcolor[HTML]{FD6864}0.3872                   & 0.003                          & 0                             \\
\multicolumn{1}{|l|}{83}                   & User                   & no i don't like that                                                    & 0                                                & 0                              & 0                             \\
\multicolumn{1}{|l|}{84}                   & Bot                    & sure finding something else                                             & 0.0001                                           & 0                              & 0                             \\
\multicolumn{1}{|l|}{85}                   & User                   & <SILENCE>                                            & 0.001                                            & 0                              & 0                             \\
\multicolumn{1}{|l|}{86}                   & Bot                    & how about this one: resto\_rome\_moderate\_italian\_3stars\_2           & \cellcolor[HTML]{FD6864}0.4834                   & \cellcolor[HTML]{FE0000}0.9968 & 0                             \\
\multicolumn{1}{|l|}{87}                   & User                   & no i don't like that                                                    & 0                                                & 0                              & 0                             \\
\multicolumn{1}{|l|}{88}                   & Bot                    & sure finding something else                                             & 0.0001                                           & 0                              & 0                             \\ \hline
\multicolumn{2}{|l|}{\textbf{User Input}}                           & \multicolumn{4}{l|}{<SILENCE>}                                                                                                                                           \\ \hline
\multicolumn{2}{|l|}{\textbf{Correct Answer}}                       & \multicolumn{4}{l|}{how about this one: resto\_rome\_moderate\_italian\_8stars\_1}                                                                                                          \\ \hline
\multicolumn{2}{|l|}{\textbf{Predicted Answer}}                     & \multicolumn{4}{l|}{how about this one: resto\_paris\_cheap\_italian\_2stars\_1}                                                                                                            \\ \hline
\end{tabular}%
}
\end{table}

\begin{table}[]
\centering
\caption{\small \textbf{Personalization task 4 (Providing information).} The model directs its attention to all the KB facts that it may need to provide but does not focus on the user profile sufficiently. Instead, it also attends to its own final utterance before the turn, which may have helped it judge the user’s gender and age instead of using the profile.  It correctly predicts that it has to display the social media information instead of the phone number for the young user, but provides the information for the wrong restaurant. \cite{BordesW16} claim that ‘embeddings mix up the information and make it hard to distinguish between different KB entities, making answering correctly very hard.’ They overcome this problem by using match type features to emphasize entities that appear in the conversation history.
\vspace{0.1in}}
\label{memnn-pt4}
\resizebox{\textwidth}{!}{%
\begin{tabular}{|l|ll|c|c|c|}
\hline
\textbf{Time} & \multicolumn{1}{l|}{\textbf{Locutor}} & \textbf{Dialog History} & \textbf{Hop \#1} & \textbf{Hop \#2} & \textbf{Hop \#3} \\ \hline
1 & \multicolumn{1}{l|}{} & male young & 0.003 & 0 & 0 \\
2 & \multicolumn{1}{l|}{User} & resto\_madrid\_cheap\_indian\_8stars\_1 R\_phone resto\_madrid\_cheap\_indian\_8stars\_1\_phone & \cellcolor[HTML]{FD6864}0.379 & \cellcolor[HTML]{FD6864}0.311 & 0.001 \\
3 & \multicolumn{1}{l|}{User} & resto\_madrid\_cheap\_indian\_8stars\_1 R\_cuisine indian & 0.001 & 0 & 0 \\
4 & \multicolumn{1}{l|}{User} & resto\_madrid\_cheap\_indian\_8stars\_1 R\_address resto\_madrid\_cheap\_indian\_8stars\_1\_address & 0.021 & 0.005 & 0 \\
5 & \multicolumn{1}{l|}{User} & resto\_madrid\_cheap\_indian\_8stars\_1 R\_location madrid & 0.003 & 0.015 & \cellcolor[HTML]{FD6864}0.342 \\
6 & \multicolumn{1}{l|}{User} & resto\_madrid\_cheap\_indian\_8stars\_1 R\_number eight & 0.002 & 0 & 0 \\
7 & \multicolumn{1}{l|}{User} & resto\_madrid\_cheap\_indian\_8stars\_1 R\_price cheap & 0.002 & 0.001 & 0 \\
8 & \multicolumn{1}{l|}{User} & resto\_madrid\_cheap\_indian\_8stars\_1 R\_rating 8 & 0.001 & 0 & 0 \\
9 & \multicolumn{1}{l|}{User} & resto\_madrid\_cheap\_indian\_8stars\_1 R\_type veg & 0.001 & 0 & 0 \\
10 & \multicolumn{1}{l|}{User} & resto\_madrid\_cheap\_indian\_8stars\_1 R\_speciality biryani & 0.002 & 0 & 0 \\
 & \multicolumn{1}{l|}{} & resto\_madrid\_cheap\_indian\_8stars\_1 R\_social\_media & \cellcolor[HTML]{FFCCC9} & \cellcolor[HTML]{FFCCC9} &  \\
\multirow{-2}{*}{11} & \multicolumn{1}{l|}{\multirow{-2}{*}{User}} & \multicolumn{1}{c|}{resto\_madrid\_cheap\_indian\_8stars\_1\_social\_media} & \multirow{-2}{*}{\cellcolor[HTML]{FFCCC9}0.084} & \multirow{-2}{*}{\cellcolor[HTML]{FFCCC9}0.069} & \multirow{-2}{*}{0.001} \\
 & \multicolumn{1}{l|}{} & resto\_madrid\_cheap\_indian\_8stars\_1 R\_parking & \cellcolor[HTML]{FD6864} & \cellcolor[HTML]{FFCCC9} &  \\
\multirow{-2}{*}{12} & \multicolumn{1}{l|}{\multirow{-2}{*}{User}} & \multicolumn{1}{c|}{resto\_madrid\_cheap\_indian\_8stars\_1\_parking} & \multirow{-2}{*}{\cellcolor[HTML]{FD6864}0.354} & \multirow{-2}{*}{\cellcolor[HTML]{FFCCC9}0.222} & \multirow{-2}{*}{0} \\
 & \multicolumn{1}{l|}{} & resto\_madrid\_cheap\_indian\_8stars\_1 R\_public\_transport &  &  &  \\
\multirow{-2}{*}{13} & \multicolumn{1}{l|}{\multirow{-2}{*}{User}} & \multicolumn{1}{c|}{resto\_madrid\_cheap\_indian\_8stars\_1\_public\_transport} & \multirow{-2}{*}{0.019} & \multirow{-2}{*}{0.001} & \multirow{-2}{*}{0} \\
14 & \multicolumn{1}{l|}{User} & hello & 0.001 & 0 & 0 \\
15 & \multicolumn{1}{l|}{Bot} & hey dude what is up & 0 & 0 & 0.004 \\
16 & \multicolumn{1}{l|}{User} & can you make a restaurant reservation at resto\_madrid\_cheap\_indian\_8stars\_1 & 0.005 & 0 & 0 \\
17 & \multicolumn{1}{l|}{Bot} & cool its done & \cellcolor[HTML]{FFCCC9}0.116 & \cellcolor[HTML]{FD6864}0.37 & \cellcolor[HTML]{FD6864}0.652 \\ \hline
\multicolumn{2}{|l|}{\textbf{User input}} & \multicolumn{4}{l|}{may i have the contact details of the restaurant} \\ \hline
\multicolumn{2}{|l|}{\textbf{Correct answer}} & \multicolumn{4}{l|}{here you go resto\_madrid\_cheap\_indian\_8stars\_1\_social\_media} \\ \hline
\multicolumn{2}{|l|}{\textbf{Predicted answer}} & \multicolumn{4}{l|}{here you go resto\_rome\_cheap\_indian\_7stars\_2\_social\_media} \\ \hline
\end{tabular}%
}
\end{table}

\begin{table}[]
\centering
\caption{\small \textbf{{Personalization task 4 with \textit{split memory} architecture.}} The model is able to single out the important user attribute (age) needed to reason over the correctly identified KB facts. However, it still mixes up embeddings of similar KB entities despite the stronger signals compared to the standard Memory Network.  
\vspace{0.1in}}
\label{memnn-sm-pt4}
\resizebox{\textwidth}{!}{%
\begin{tabular}{lll|c|c|c|}
\cline{3-6}
                                           & \multicolumn{1}{l|}{}                       & \multicolumn{1}{c|}{\textbf{Profile}}                                            & \textbf{Hop \#1}                                & \textbf{Hop \#2}              & \textbf{Hop \#3}              \\ \cline{3-6} 
                                           & \multicolumn{1}{l|}{}                       & \multicolumn{1}{c|}{male}                                                        & \cellcolor[HTML]{FD6864}0.431                   & 0.091                         & 0.002                         \\
                                           & \multicolumn{1}{l|}{}                       & \multicolumn{1}{c|}{young}                                                       & \cellcolor[HTML]{FD6864}0.569                   & \cellcolor[HTML]{FE0000}0.909 & \cellcolor[HTML]{FE0000}0.998 \\ \cline{1-3}
\multicolumn{1}{|l|}{\textbf{Time}}        & \multicolumn{1}{l|}{\textbf{Locutor}}       & \textbf{Dialog History}                                                          &                                                 &                               &                               \\ \cline{1-3}
\multicolumn{1}{|l|}{}                     & \multicolumn{1}{l|}{}                       & resto\_madrid\_cheap\_indian\_8stars\_1 R\_phone                                 &                                                 &                               &                               \\
\multicolumn{1}{|l|}{\multirow{-2}{*}{1}}  & \multicolumn{1}{l|}{\multirow{-2}{*}{User}} & \multicolumn{1}{c|}{resto\_madrid\_cheap\_indian\_8stars\_1\_phone}              & \multirow{-2}{*}{0.001}                         & \multirow{-2}{*}{0}    & \multirow{-2}{*}{0}    \\
\multicolumn{1}{|l|}{2}                    & \multicolumn{1}{l|}{User}                   & resto\_madrid\_cheap\_indian\_8stars\_1 R\_cuisine indian                        & 0                                               & 0                      & 0                      \\
\multicolumn{1}{|l|}{}                     & \multicolumn{1}{l|}{}                       & resto\_madrid\_cheap\_indian\_8stars\_1 R\_address                               &                                                 &                               &                               \\
\multicolumn{1}{|l|}{\multirow{-2}{*}{3}}  & \multicolumn{1}{l|}{\multirow{-2}{*}{User}} & \multicolumn{1}{c|}{resto\_madrid\_cheap\_indian\_8stars\_1\_address}            & \multirow{-2}{*}{0}                             & \multirow{-2}{*}{0}    & \multirow{-2}{*}{0}    \\
\multicolumn{1}{|l|}{4}                    & \multicolumn{1}{l|}{User}                   & resto\_madrid\_cheap\_indian\_8stars\_1 R\_location madrid                       & 0                                               & 0                      & 0                      \\
\multicolumn{1}{|l|}{5}                    & \multicolumn{1}{l|}{User}                   & resto\_madrid\_cheap\_indian\_8stars\_1 R\_number eight                          & 0                                               & 0                      & 0                      \\
\multicolumn{1}{|l|}{6}                    & \multicolumn{1}{l|}{User}                   & resto\_madrid\_cheap\_indian\_8stars\_1 R\_price cheap                           & 0                                               & 0                      & 0                      \\
\multicolumn{1}{|l|}{7}                    & \multicolumn{1}{l|}{User}                   & resto\_madrid\_cheap\_indian\_8stars\_1 R\_rating 8                              & 0                                               & 0                      & 0                      \\
\multicolumn{1}{|l|}{8}                    & \multicolumn{1}{l|}{User}                   & resto\_madrid\_cheap\_indian\_8stars\_1 R\_type veg                              & 0                                               & 0                      & 0                      \\
\multicolumn{1}{|l|}{9}                    & \multicolumn{1}{l|}{User}                   & resto\_madrid\_cheap\_indian\_8stars\_1 R\_speciality biryani                    & 0                                               & 0                      & 0                      \\
\multicolumn{1}{|l|}{}                     & \multicolumn{1}{l|}{}                       & resto\_madrid\_cheap\_indian\_8stars\_1 R\_social\_media                         & \cellcolor[HTML]{FD6864}                        &                               &                               \\
\multicolumn{1}{|l|}{\multirow{-2}{*}{10}} & \multicolumn{1}{l|}{\multirow{-2}{*}{User}} & \multicolumn{1}{c|}{resto\_madrid\_cheap\_indian\_8stars\_1\_social\_media}      & \multirow{-2}{*}{\cellcolor[HTML]{FD6864}0.21}  & \multirow{-2}{*}{0}    & \multirow{-2}{*}{0}    \\
\multicolumn{1}{|l|}{}                     & \multicolumn{1}{l|}{}                       & resto\_madrid\_cheap\_indian\_8stars\_1 R\_parking                               & \cellcolor[HTML]{FD6864}                        &                               &                               \\
\multicolumn{1}{|l|}{\multirow{-2}{*}{11}} & \multicolumn{1}{l|}{\multirow{-2}{*}{User}} & \multicolumn{1}{c|}{resto\_madrid\_cheap\_indian\_8stars\_1\_parking}            & \multirow{-2}{*}{\cellcolor[HTML]{FD6864}0.385} & \multirow{-2}{*}{0}    & \multirow{-2}{*}{0}    \\
\multicolumn{1}{|l|}{}                     & \multicolumn{1}{l|}{}                       & resto\_madrid\_cheap\_indian\_8stars\_1 R\_public\_transport                     & \cellcolor[HTML]{FFCCC9}                        &                               &                               \\
\multicolumn{1}{|l|}{\multirow{-2}{*}{12}} & \multicolumn{1}{l|}{\multirow{-2}{*}{User}} & \multicolumn{1}{c|}{resto\_madrid\_cheap\_indian\_8stars\_1\_public\_transport}  & \multirow{-2}{*}{\cellcolor[HTML]{FFCCC9}0.066} & \multirow{-2}{*}{0}    & \multirow{-2}{*}{0}    \\
\multicolumn{1}{|l|}{13}                   & \multicolumn{1}{l|}{User}                   & hello                                                                            & 0                                               & 0                      & 0                      \\
\multicolumn{1}{|l|}{14}                   & \multicolumn{1}{l|}{Bot}                    & hey dude what is up                                                              & \cellcolor[HTML]{FD6864}0.309                   & \cellcolor[HTML]{FE0000}0.999 & \cellcolor[HTML]{FE0000}0.999 \\
\multicolumn{1}{|l|}{15}                   & \multicolumn{1}{l|}{User}                   & can you make a restaurant reservation at resto\_madrid\_cheap\_indian\_8stars\_1 & \cellcolor[HTML]{FFCCC9}0.021                   & 0                      & 0                      \\
\multicolumn{1}{|l|}{16}                   & \multicolumn{1}{l|}{Bot}                    & cool its done                                                                    & 0.005                                           & 0                      & 0                      \\ \hline
\multicolumn{2}{|l|}{\textbf{User Input}}                                                & \multicolumn{4}{l|}{may i have the contact details of the restaurant}                                                                                                                              \\ \hline
\multicolumn{2}{|l|}{\textbf{Correct Answer}}                                            & \multicolumn{4}{l|}{here you go resto\_madrid\_cheap\_indian\_8stars\_1\_social\_media}                                                                                                            \\ \hline
\multicolumn{2}{|l|}{\textbf{Predicted Answer}}                                          & \multicolumn{4}{l|}{here you go resto\_paris\_cheap\_indian\_2stars\_2\_social\_media}                                                                                                             \\ \hline
\end{tabular}%
}
\end{table}

\begin{table}[]
\centering
\caption{\small \textbf{Predictions of multi-profile model versus profile-specific model.} For the chosen profile (female, middle-aged), the multi-profile model attends to the user’s profile, greeting and incomplete inquiry to modify its speech style and ask for the missing field. The profile-specific model does not need to perform such personalization, and has a narrower focus on the inquiry. 
\vspace{0.1in}}
\label{memnn-mtl}
\resizebox{\textwidth}{!}{%
\begin{tabular}{|l|l|l|c|c|}
\hline
 &  &  & \textbf{Multi-profile} & \textbf{Profile-specific} \\
\multirow{-2}{*}{\textbf{Time}} & \multirow{-2}{*}{\textbf{Locutor}} & \multirow{-2}{*}{\textbf{Dialog History}} & \textbf{model} & \textbf{model} \\ \hline
1 &  & female middle-aged non-veg pizza & \cellcolor[HTML]{FFCCC9}0.044 & 0 \\
2 & User & good morning & \cellcolor[HTML]{FD6864}0.397 & 0 \\
3 & Bot & hello maam how can i help you & 0.005 & 0.012 \\
4 & User & can you make a restaurant reservation for two people in bombay with italian cuisine & \cellcolor[HTML]{FD6864}0.533 & \cellcolor[HTML]{FE0000}0.987 \\
5 & Bot & give me a second for processing the reservation & 0 & 0 \\ \hline
\multicolumn{2}{|l|}{\textbf{User input}} & \multicolumn{3}{l|}{<SILENCE>} \\ \hline
\multicolumn{2}{|l|}{\textbf{Correct answer}} & \multicolumn{3}{l|}{which price range are you looking for} \\ \hline
\multicolumn{2}{|l|}{\textbf{Predicted answer}} & \multicolumn{3}{l|}{which price range are you looking for (for both models)} \\ \hline
\end{tabular}%
}
\end{table}

\begin{table}[]
\centering
\caption{\small \textbf{Hyperparameters for Supervised Embeddings.} If Use History is True, conversation history is added to the last user utterance to create the input. If False, only the last utterance is used as input.
\vspace{0.1in}}
\label{se-hp}
\begin{tabular}{lccccc}
\toprule
\multicolumn{1}{l}{\multirow{2}{*}{\textbf{Task}}} & \textbf{Learning} & \multirow{2}{*}{\textbf{Margin}} & \textbf{Embedding} & \textbf{Negative} & \textbf{Use} \\
\multicolumn{1}{l}{} & \textbf{Rate} &  & \textbf{Dimension} & \textbf{Candidates} & \textbf{History} \\ \midrule
PT1 & 0.01 & 0.01 & 32 & 100 & True \\
PT2 & 0.01 & 0.01 & 128 & 100 & False \\
PT3 & 0.01 & 0.1 & 128 & 1000 & False \\
PT4 & 0.001 & 0.1 & 128 & 1000 & False \\
PT5 & 0.01 & 0.01 & 32 & 100 & True \\ \bottomrule
\end{tabular}
\end{table}

\begin{table}[]
\centering
\caption{\small \textbf{Hyperparameters for Memory Networks.} Models with the \textit{Split memory} architecture were also trained using the same hyperparameters.
\vspace{0.1in}}
\label{se-memnn}
\begin{tabular}{lccccc}
\toprule
\multicolumn{1}{l}{\multirow{2}{*}{\textbf{Task}}} & \textbf{Learning} & \multirow{2}{*}{\textbf{Margin}} & \textbf{Embedding} & \textbf{Negative} & \textbf{Number} \\
\multicolumn{1}{l}{} & \textbf{Rate} &  & \textbf{Dimension} & \textbf{Candidates} & \textbf{of Hops} \\ \midrule
PT1 & 0.001 & 0.01 & 20 & 100 & 1 \\
PT2 & 0.001 & 0.01 & 20 & 100 & 1 \\
PT3 & 0.001 & 0.01 & 20 & 100 & 3 \\
PT4 & 0.001 & 0.01 & 20 & 100 & 3 \\
PT5 & 0.001 & 0.01 & 20 & 100 & 3 \\ \bottomrule
\end{tabular}
\end{table}

\end{document}